\documentclass[letterpaper]{article} 
\usepackage{aaai23}  
\usepackage{times}  
\usepackage{helvet}  
\usepackage{courier}  
\usepackage[hyphens]{url}  
\usepackage{graphicx} 
\usepackage{natbib}  
\usepackage{caption} 
\usepackage{algorithm}
\usepackage{algorithmic}
\usepackage{amsfonts}
\usepackage{amsmath}
\usepackage{tikz}
\usepackage{subcaption}
\usepackage[export]{adjustbox}
\usepackage{newfloat}
\usepackage{listings}
\DeclareCaptionStyle{ruled}{labelfont=normalfont,labelsep=colon,strut=off} 
\floatstyle{ruled}
\newfloat{listing}{tb}{lst}{}
\floatname{listing}{Listing}
\title{Automaton-Guided Curriculum Generation for Reinforcement Learning Agents}
\author{
    Yash Shukla\textsuperscript{\rm 1},
    Abhishek Kulkarni\textsuperscript{\rm 2},
    Robert Wright\textsuperscript{\rm 3},
    Alvaro Velasquez\textsuperscript{\rm 4},
    Jivko Sinapov\textsuperscript{\rm 1}
}
\affiliations{

    \textsuperscript{\rm 1}Tufts University \\
    \textsuperscript{\rm 2}University of Florida \\
    \textsuperscript{\rm 3}Georgia Tech Research Institute \\ 
    \textsuperscript{\rm 4}University of Colorado Boulder 

    \{yash.shukla, jivko.sinapov\}@tufts.edu,
    a.kulkarni2@ufl.edu,
    robert.wright@gtri.gatech.edu,
    alvaro.velasquez@colorado.edu
}

\usepackage{bibentry}

\begin{document}

\maketitle

\begin{abstract}

Despite advances in Reinforcement Learning, many sequential decision making tasks remain prohibitively expensive and impractical to learn. Recently, approaches that automatically generate reward functions from logical task specifications have been proposed to mitigate this issue; however, they scale poorly on long-horizon tasks (i.e., tasks where the agent needs to perform a series of correct actions to reach the goal state, considering future transitions while choosing an action). Employing a curriculum (a sequence of increasingly complex tasks) further improves the learning speed of the agent by sequencing intermediate tasks suited to the learning capacity of the agent. However, generating curricula from the logical specification still remains an unsolved problem. To this end, we propose \emph{AGCL}, Automaton-guided Curriculum Learning, a novel method for automatically generating curricula for the target task in the form of Directed Acyclic Graphs (DAGs). \emph{AGCL} encodes the specification in the form of a deterministic finite automaton (DFA), and then uses the DFA along with the Object-Oriented MDP (OOMDP) representation to generate a curriculum as a DAG, where the vertices correspond to tasks, and edges correspond to the direction of knowledge transfer. Experiments in gridworld and physics-based simulated robotics domains show that the curricula produced by \emph{AGCL} achieve improved time-to-threshold performance on a complex sequential decision-making problem relative to state-of-the-art curriculum learning (e.g, teacher-student, self-play) and automaton-guided reinforcement learning baselines (e.g, Q-Learning for Reward Machines). Further, we demonstrate that \emph{AGCL} performs well even in the presence of noise in the task's OOMDP description, and also when distractor objects are present that are not modeled in the logical specification of the tasks' objectives.

\end{abstract}

\section{Introduction}

Deep reinforcement learning utilizes neural networks to solve complex tasks ranging from Atari games to robot manipulation~\cite{gao2021efficiently,karpathy2012curriculum}. Despite these advances, many tasks remain prohibitively expensive to learn from scratch, requiring large number of interactions. This problem worsens in long-horizon tasks where sparse reward settings and poor goal representations make the task difficult to solve. In many practical scenarios, the objective for a task is known before commencing learning the task, and thus it can be captured using high-level specifications that have an equivalent automaton representation~\cite{velasquez2021dynamic, jothimurugan2021compositional}. This formulation allows decomposing the task into sub-goals, where each sub-goal can be learned by a Reinforcement Learning (RL) agent. This approach of automaton-guided RL is beneficial in scenarios where the high-level task objective is known beforehand and can be expressed using an automaton, but the low-level transition dynamics are unavailable and hence motion and symbolic planners cannot be reliably used. Recently, many approaches have made use of richer language representations (such as finite-trace Linear Temporal Logic (LTL$_f$)~\cite{de2013linear}) to incorporate history in the task Markovian Decision Processes (MDPs). This technique helps the RL agent to shape reward effectively to achieve the goal in sparse reward settings, eliminating the need to rely on human guidance for shaping the reward function~\cite{camacho2018non,icarte2018using,velasquez2021dynamic}. 
However, reward shaping falls short in complex sequential-decision making tasks as it does not generate simpler sub-tasks suited to the current knowledge of the agent, and incentivizes policies to reach local optima. In this work, we extend automaton-guided RL beyond reward shaping, by deriving a curriculum for the target objective. Curriculum Learning (CL) intends to optimize the order in which an agent attempts tasks, increasing the expected return while reducing training time for complex tasks~\cite{narvekar2020curriculum,foglino2019optimization}. Nevertheless, automaton-guided curriculum generation still remains an unsolved problem in literature.
\begin{figure}[t]
    \centering
    \includegraphics[width=0.45\textwidth,height=0.35\textwidth]{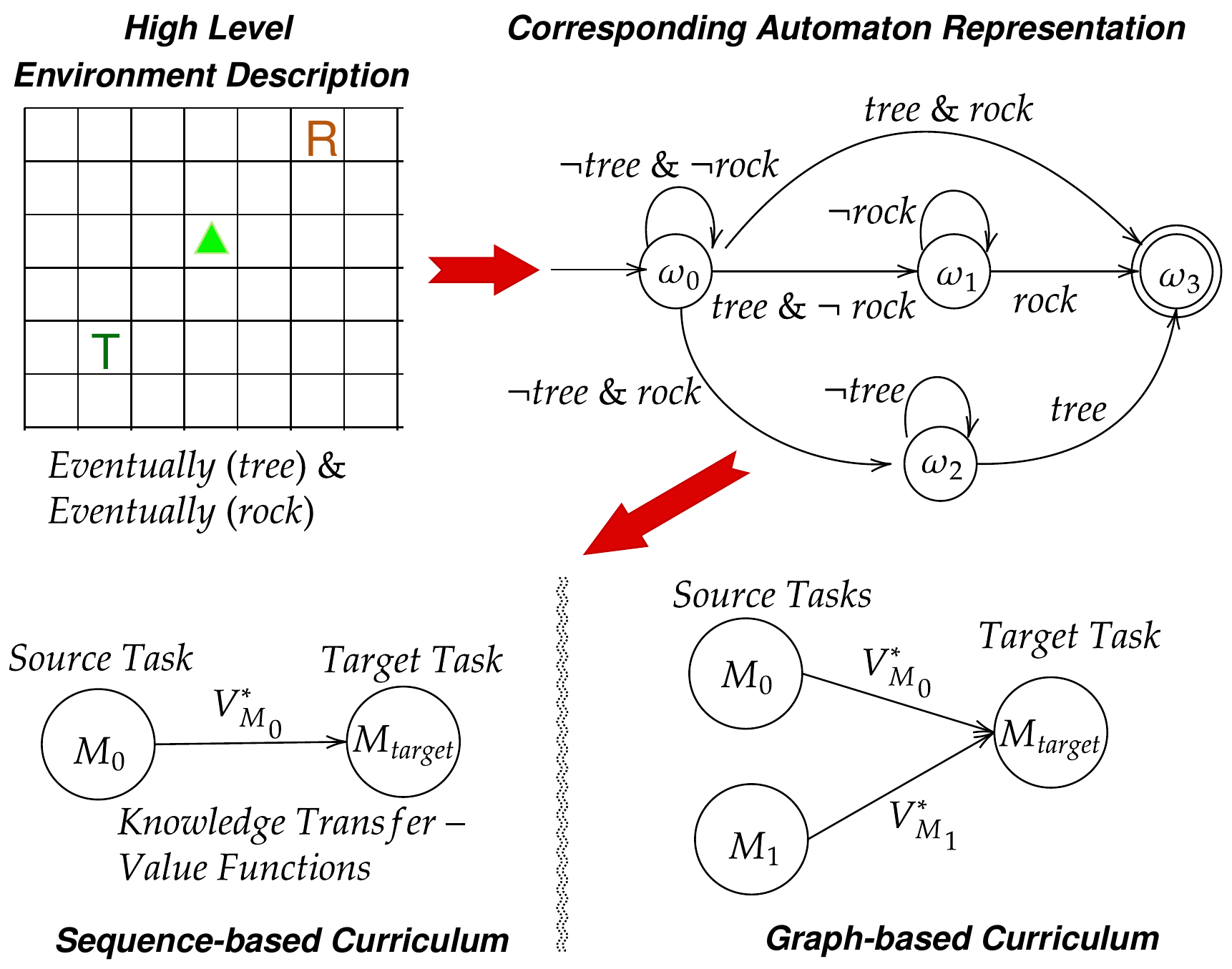}
    \caption{Overview of the automaton-guided curriculum generation procedure. Given the high-level task description, we use the equivalent automaton representation to come up with a sequence-based and a  graph-based curriculum.
    }
    \label{fig:overview}
\end{figure}


We explore representing the high-level task objective using finite-trace Linear Temporal Logic (LTL$_f$) formulas~\cite{de2013linear} which can be equivalently represented using Deterministic Finite Automaton (DFA). Using the structure of the DFA, we generate a curriculum for the target objective. While the DFA provides a graphical representation of the sequence in which the sub-goals must be achieved, it does not specify the individual sub-tasks of the curriculum. Thus, generating a curriculum is non-trivial as it requires the agent to reason over multiple potential curriculum environment configurations for the same sub-goal objective. Given a high-level task objective in the form of a DFA, we propose \emph{AGCL}, Automaton-guided Curriculum Learning (Fig.~\ref{fig:overview}), that generates two types of curricula: $1)$ A sequence of sub-tasks that increase in order of difficulty for the agent, and $2)$ A directed acyclic graph (DAG), where agents can transfer knowledge concurrently learned in multiple source tasks to learn a common target task\footnote{An in-depth analysis of Fig. 1 is provided in Section 3}. Unlike previous graph-based curriculum approaches~\cite{svetlik2017automatic, silva2018object}, ours does not require access to all possible state configurations of the environment. Further, generating the curriculum does not require additional interactions with the environment, and learning through the curriculum yields quicker convergence to a desired performance compared to learning from scratch, automaton-guided reward shaping baselines - GSRS~\cite{camacho2018non}, QRM~\cite{icarte2018using}, and curriculum learning baselines - Teacher-Student~\cite{DBLP:journals/tnn/MatiisenOCS20} and self-play~\cite{sukhbaatar2018intrinsic}. 
 
We further perform an extensive evaluation on a set of challenging robotic navigation and manipulation tasks and demonstrate that \emph{AGCL} reduces the number of interactions with the target environment by orders-of-magnitude when compared to state-of-the-art curriculum learning and automaton-guided reinforcement learning baselines.      
 
\section{Related Work} 

\textbf{Curriculum Learning} (CL) in RL has been studied for games \cite{gao2021efficiently}, robotic tasks \cite{karpathy2012curriculum}, and self-driving cars \cite{qiao2018automatically}.  A survey on CL~\cite{narvekar2020curriculum} summarizes the three main elements of CL as task generation, task sequencing, and knowledge transfer. Task generation builds a set of source tasks given the target task~\cite{kurkcu2020autonomous}.
Task sequencing optimizes the task sequence to enhance learning in the target task \cite{narvekar2017autonomous, DBLP:journals/tnn/MatiisenOCS20}, and knowledge transfer determines what information must be transferred from the source to the target to promote effective learning~ \cite{da2019survey, taylor2009transfer}.
Existing graph-based CL methods~\cite{svetlik2017automatic, silva2018object} assume knowledge of the set of states in the task, and thus cannot be used for complex tasks that require function approximation. Metaheuristic search methods serve as a tool to evaluate the CL frameworks~\cite{foglino2019optimization}. In most methods, optimizing a curriculum is still computationally expensive and sometimes takes more interactions compared to learning from scratch. Our proposed framework addresses this concern by utilizing high-level logical specification to derive an optimized graphical representation of a curriculum, eliminating the sunk cost of interactions required to optimize the curriculum. \\
\textbf{Automaton-guided RL} approaches specify tasks using temporal logic-based high-level language specifications~\cite{toro2018teaching,velasquez2021dynamic,jiang2021temporal}. Most approaches generate a dense reward function to speed up learning. Learning separate policies for sub-goals of a task helps abstract knowledge which can be used in new tasks~\cite{icarte2018using}. Another technique is to shape the reward inversely proportional to the distance from the accepting node in the automaton~\cite{camacho2018non} however, it leads to inefficient reward settings. Augmenting the reward function with a dynamic Monte Carlo Tree Search helps mitigate this problem~\cite{velasquez2021dynamic}. DIRL interleaves high-level planning with RL to learn a policy for each edge, overcoming challenges introduced by poor representations~\cite{jothimurugan2021compositional}. However, this approach becomes inefficient when there are multiple paths to the target in the graph, and other reward shaping techniques prove inefficient compared to curriculum learning~\cite{pocius2018comparing}. In this work, we use the underlying DFA to develop a graphical representation of the curriculum, and implement a transfer learning mechanism compatible with a function-approximator, outperforming state-of-the-art baselines.
 
\section{Theoretical Framework}

\subsection{Markov Decision Processes}
An episodic Markov Decision Process (MDP) $M$ is defined as a tuple $(\mathcal{S},\! \mathcal{A},\! \mathcal{P},\! R\!, \mathcal{S}_0, \mathcal{S}_f, \gamma, L)$, where $\mathcal{S}$ is the set of states, $\mathcal{A}$ is the set of actions, $\!\mathcal{P}(s'\!|s,a)\!$ is the transition function,~   
$\!R:\! \mathcal{S}\! \times\! A \!\times \mathcal{S}\! \rightarrow \!\mathbb{R}$ is the reward function, $\mathcal{S}_0$ and $\mathcal{S}_f$ are the sets of starting and final states respectively, and $\gamma\!\in\! [0,1]$ is the discount factor. At each timestep $t$, the agent observes a state $s$ and performs an action $a$ given by its policy function $\pi_\theta(a|s)$, with parameters $\theta$. The agent's goal is to learn an \emph{optimal policy $\pi^*\!$}, maximizing its discounted return $G_0 = \sum^{K}_{k = 0}\!\gamma^k\! r(s'_k,a_k,s_k) $ until the end of the episode at timestep $K$. The labeling function $L: \mathcal{S} \rightarrow 2^{AP}$ maps a state in the MDP to a set of atomic propositions that hold true for that state. For example, in Fig.~\ref{fig:overview} the set of atomic propositions, $AP = \{tree,rock\}$.

\begin{figure*}
    \begin{center}
        \begin{tabular}{ c | c }
            {\includegraphics[width=0.38\textwidth, height=0.18\textwidth,valign=c]{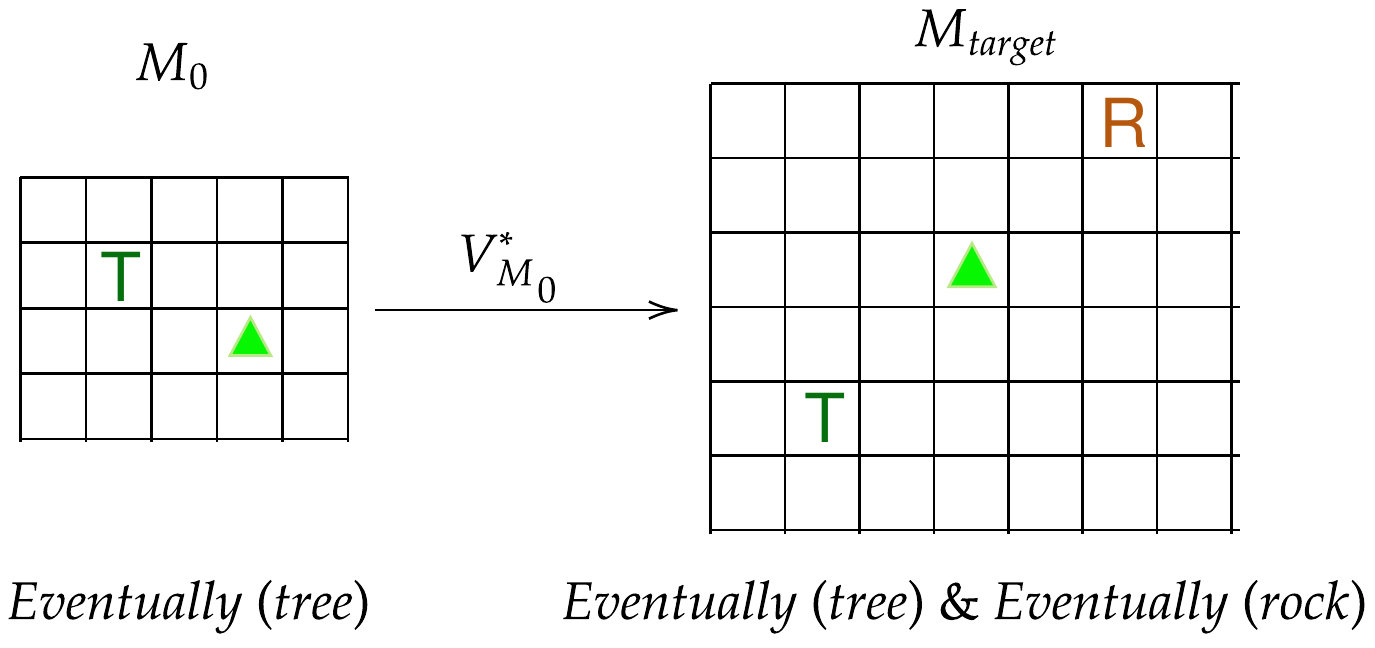}} &
            {\includegraphics[width=0.45\textwidth,height=0.25\textwidth,valign=c]{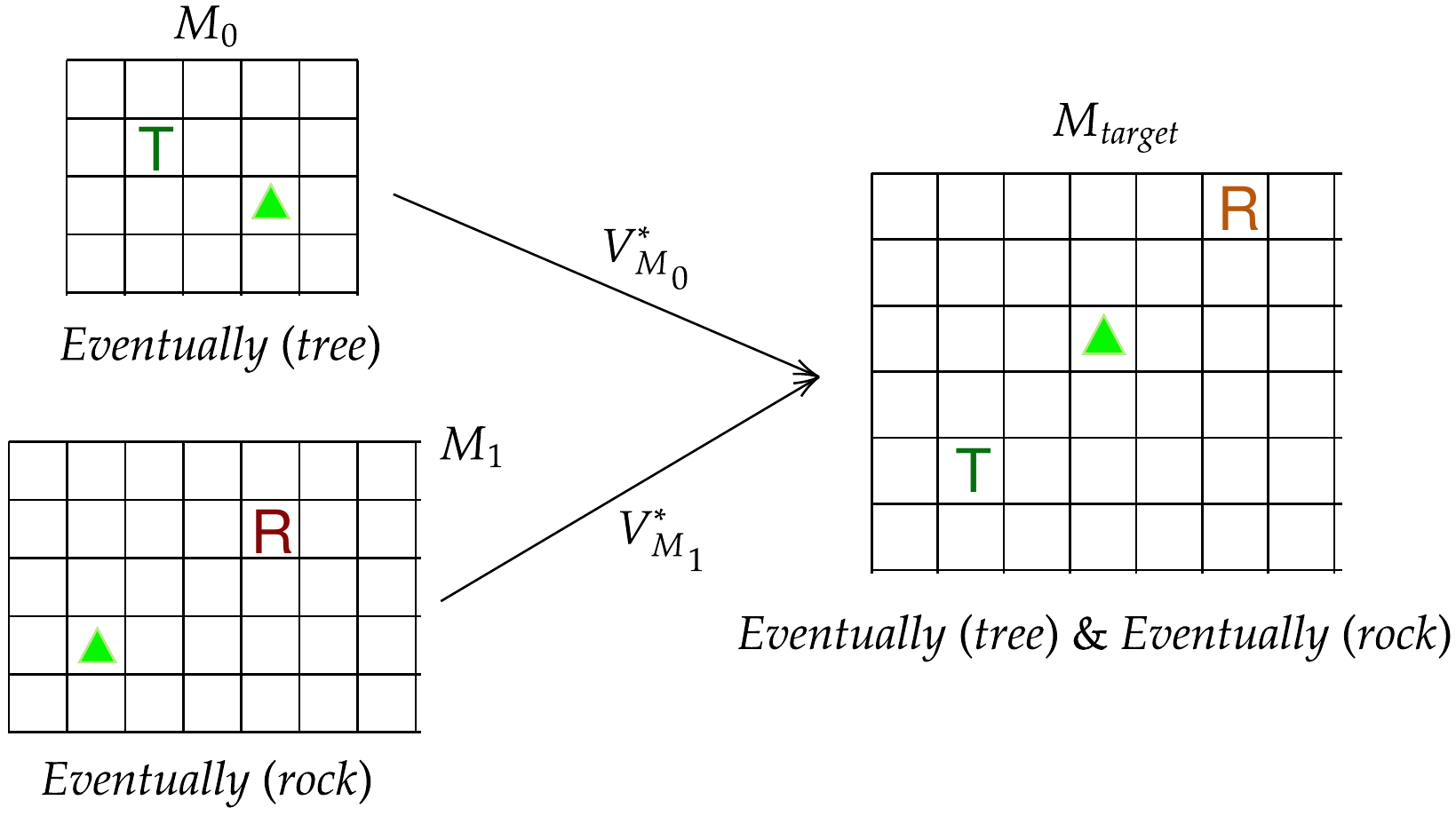}}
            \\
            {{(a) Sequence-based curriculum}} &
            {{(b) Graph-based curriculum}} 
        \end{tabular}
    \end{center}
    \caption{Examples of sequence-based (left) and graph-based (right) curricula. \label{fig:curriculum_examples}}
\end{figure*}
\subsection{Curriculum}
Let $\mathcal{M}$ be a set of tasks, where $M_i=  (\mathcal{S}_i,\mathcal{A}_i,P_i,R_i, \mathcal{S}_{0,i}, \mathcal{S}_{f,i}, \gamma_i, L_i) $ is  a task  in $\mathcal{M}$. Let $\mathcal{D}^{M_i}$ be the set of  samples  associated with task $\!\!M_i:\!\! \mathcal{D}^{M_i}\!= \!\!\{ (s,a,r,s')| s \in \mathcal{S}_i , a \in \mathcal{A}_i, s' \sim \mathcal{P}_i(\cdot|s,a), r \leftarrow R_i(s,a,s')\}$. A curriculum $T=  (\mathcal{V},\mathcal{E},g,\mathcal{M})$ is  a  directed  acyclic  graph,  where $\mathcal{V}$ is the set of vertices, $\mathcal{E} \subseteq \{(x,y)|(x,y) \in V \times V \wedge x \neq y\}$ is the set of directed edges, and $g: \mathcal{V} \rightarrow \{\mathcal{D}^{M_i} | M_i \in \mathcal{M}\}$ is a function that associates vertices to samples of a single task in $\mathcal{M}$. A  directed  edge $(v_j,v_k)$ indicates that samples associated with $v_j \in \mathcal{V}$ should be trained on before samples associated with $v_k \in \mathcal{V}$. All paths terminate on a single sink node $v_t \in \mathcal{V}$\footnote{We find curriculum for single target task. Curriculum design for multiple target tasks is beyond the scope of this paper.}.


A graph-based curriculum is a general case of a curriculum, where the indegree and outdegree of each vertex $v\in \mathcal{V}$ can be greater than one, and there can be multiple source nodes but only one sink node. A sequence-based curriculum is a special case of a curriculum, where the indegree and outdegree of each vertex $v\in \mathcal{V}$ is at most 1, and there is only one source node and one sink node as shown in Fig.~\ref{fig:overview}. Examples of a sequence-based curriculum and a graph-based curriculum are shown in Fig.~\ref{fig:curriculum_examples}(a) Fig.~\ref{fig:curriculum_examples}(b) respectively.

\subsection{Object Oriented MDP} \label{sec:oomdp}
The Object-Oriented MDP (OOMDP) representation abstracts the task description, as to intuitively generalize the elements in the environment and their properties~\cite{diuk2008object}. In OOMDPs, the task space is abstracted by a set of classes $\mathcal{C}$ where each class $C_i \!\in \!\mathcal{C}$ has a parameter set $Par(C_i)\! = \!\{C_i.p_1,\! \ldots,\! C_i.p_{|p|}\}$. Each parameter $C_i.p_i$ has a range of values, given by $Range(C_i.p_i)$. At any given time, an environment consists of a set of objects $\mathcal{O}\! = \!\{o_1,\! \ldots,\! o_{|o|} \}$, where each object $o_i$ is an instance of a class $C_i$ and is defined by the parameters for that class $o_i = Par(C_i(o_i))$. For a task, the OOMDP state $s_{oo}$ is given by the union of all object states $s_{oo}\!\! = \!\! \cup_{o_i\in \mathcal{O}}  State(o_i)$, where each object state is the value of the parameters of that object: $State(o_i)\!\!\! =\!\!\! \{o_i.p_1, \ldots, o_i.p_{|p|} \}$. Since the OOMDP representation is a high-level abstraction of the task space, we assume a many-to-one mapping between a subset of MDP states ($\mathcal{S}_{sub}$) and the OOMDP state ($s_{oo}$); $w \!:\! \mathcal{S}_{sub}\! \rightarrow\! s_{oo}$.
For the task in Fig.~\ref{fig:overview}, the OOMDP description consists of classes $\mathcal{C} = \{world\_size, trees, rocks\}$, with the parameters: $\mathcal{C}(world\_size) = \{width, height\}$  (width and height of the grid), $\mathcal{C}(trees) = \{trees_{env}, trees_{inv}\}$ (number of trees in the environment and inventory of the agent), and similarly, 
$\mathcal{C}(rocks) = \{rocks_{env}, rocks_{inv}\}$. In a task, classes have objects whose parameters are assigned values within a range. Example: $Par(world\_size[height])) = [5,10]$ denotes that the height of the grid can have any value within the range $[5,10]$. The MDP transition $s\!\! \xrightarrow{a}\!\! s'$ and the OOMDP transition $s_{oo}
\!\!\xrightarrow{a}\!\! s'_{oo}$ occur synchronously.

\subsection{Linear Temporal Logic and Deterministic Finite Automata (DFA)}

We define the high-level specification of our task using finite-trace Linear Temporal Logic (LTL$_f$) formulas~\cite{de2013linear}. LTL$_f$ allows us to succinctly express complex temporal tasks such as reachability, safety, recurrence, persistence or a combination of these. Most importantly, the language of any LTL$_f$ formula can equivalently be represented as a Deterministic Finite Automaton (DFA). 

Given a set of atomic propositions, $AP$, a formula in LTL$_f$ is constructed inductively using the operations $p, \neg \phi, \phi_1 \vee \phi_2$, \textbf{X}$\phi$, \textbf{G}$\phi$, \textbf{F}$\phi$,$\phi_1$\textbf{U}$\phi_2$ where $p \in AP$ and $\phi, \phi_1$ and $\phi_2$ are LTL$_f$ formulas. The operator $\neg$ denotes negation, $\vee$ denotes disjunction, and the operators \textbf{X}, \textbf{G}, \textbf{F}, \textbf{U} denote \textit{Next}, \textit{Always}, \textit{Eventually}, and \textit{Until} respectively.
 An LTL$_f$ formula can be translated into a DFA $\mathcal{U} = \{\Omega, \omega_0, \Sigma, \delta, F \}$, where $\Omega$ is the set of nodes with initial node $\omega_0 \in \Omega$, $\Sigma = 2^{AP}$ is an alphabet defined over a set of atomic propositions AP, $\delta : \Omega \times \Sigma \rightarrow \Omega$ is a deterministic transition function, and $F\subseteq \Omega$ is the set of accepting nodes. Given any two states $\omega, \omega' \in \Omega$ and a symbol $\sigma \in \Sigma$, we denote the transition $\omega' = \delta(\omega, \sigma)$ as $\omega \xrightarrow{\sigma} \omega'$. A path $\zeta = \omega_0 \omega_1 \ldots \omega_n$ is a finite-length sequence of states such that, for any $i = 0 \ldots n-1$, we have $\omega_{i+1} =
\delta(\omega_i, \sigma_i)$. The occurrence operator, $Occ(\zeta)$, defined as $\{\omega \in \Omega | \omega \mbox{ appears in } \zeta\}$, returns the set of nodes in the path $\zeta$.
 
The set of atomic propositions in $AP$ is shared within the MDP and the DFA. The labeled MDP and the automaton (DFA) operate synchronously. Specifically, a transition $s \xrightarrow{a} s'$ in the labeled MDP triggers a transition $\omega
\xrightarrow{L(s')} \omega'$ in the DFA. Therefore, the high-level specification is achieved in the labeled MDP if and only if a final state is reached in the DFA. We define a one-to-many relationship using the function $P: \Omega \rightarrow \mathcal{M}$, that given a node $\omega \in \Omega$ in the DFA, returns the set of tasks that can reach the node $\omega$ from the start node $\omega_0$\footnote{Details on function $P$ in Appendix Section A. \\ https://github.com/tufts-ai-robotics-group/Automaton-guided-CL}. The equivalent DFA representation of the LTL$_f$ objective \textbf{F}$(tree)\wedge$\textbf{F}$(rock)$ is shown in Fig.~\ref{fig:overview}.

\subsection{Running Example}
Consider the environment description shown in Fig.~\ref{fig:overview}. We start with the target task MDP $M_{target}$ and its OOMDP description (described in Sec.~\ref{sec:oomdp}) along with the LTL$_f$ formula for the task (\textbf{F}$(tree) \wedge$ \textbf{F}$(rock)$) which corresponds to the automaton shown in Fig.~\ref{fig:overview}. When the agent follows a  policy $\pi$ that collects a tree followed by a rock, it satisfies the path $\zeta: \omega_0 \xrightarrow{tree} \omega_1 \xrightarrow{rock} \omega_3$. Since $\omega_3$ is in the set of accepting nodes $F$, the policy produces a trajectory in the MDP that satisfies the LTL$_f$ objective. Generating curriculum given the DFA involves reasoning over the trace paths as well as the set of MDPs that can generate trajectories to reach a node in the DFA. For example, to have a trajectory (of length $>0$) that reaches the node $\omega_1$, the agent must have a \emph{tree} in its inventory by the end of the episode. To achieve this, at least one \emph{tree} must be present in the environment when the episode starts for the agent to collect. Using this information, a potential source task of the curriculum is generated by varying the parameter values for the objects within the range specified in the OOMDP representation. Thus, a query to the function $P(\omega_1)$ will generate the set of MDPs $\mathcal{M}_1$ that have at least $1$ \emph{tree} in the initial task configuration, while varying other object parameters. The set of initial and final OOMDP states for $\mathcal{M}_1$ will be given by $w(P(\omega_1)(\mathcal{S}_0))$ and $w(P(\omega_1)(\mathcal{S}_f))$ respectively as the function $w$ maps the MDP state to an OOMDP state. An example of a sequence-based curriculum for the path $\zeta: \omega_0 \xrightarrow{tree} \omega_1 \xrightarrow{rock} \omega_3$ is shown in Fig.~\ref{fig:curriculum_examples}, where in the first task $M_0$, the agent learns to collect a \emph{tree} in a smaller environment, while transferring the knowledge to the final task. For the graph-based curriculum in Fig.~\ref{fig:curriculum_examples}, two agents simultaneously learn to collect a \emph{tree} and a \emph{rock} separately, and transfer their knowledge to learn the target task. The curriculum for the agent that learns to collect a \emph{tree} in $M_0$ is derived from the set of MDPs produced using the path $\zeta: \omega_0 \xrightarrow{tree} \omega_1 \xrightarrow{rock} \omega_3$, whereas the curriculum for the agent that learns to collect a \emph{rock} in $M_1$ is derived using the path $\zeta: \omega_0 \xrightarrow{rock} \omega_2 \xrightarrow{tree} \omega_3$. The agent's trajectory in the final target task $M_{target}$ can follow any of the paths in the DFA. The knowledge from the source tasks in the curriculum guides the agent in the final target task to make decisions based on its prior experiences. In the following sections, we discuss our automaton-guided curriculum generation approach.

\subsection{Problem Formulation}
CL aims to generate a curriculum, such that the agent's convergence or its time-to-threshold performance on the final target task $(M_{target})$ improves relative to learning from scratch. Time-to-threshold ($\Delta$) metric computes how much faster an agent can learn a policy that achieves expected return $G \geq \delta$ on the target task if it learns through a curriculum, as opposed to learning from another approach~\cite{foglino2019optimization}. Here $\delta$ is desired performance threshold, in success rate or episodic reward. An optimal curriculum would converge to the expected return value in the target task the quickest. Formally:
\[\small T^* := \operatorname*{argmin}_{T \in \mathcal{T}} [a_{\delta}^{M_{target}} + \sum_{M_i \in T} a^{M_i}] \]
where $\mathcal{T}$ is the set of all curricula, $a_{\delta}^{M_{target}}$ is the number of actions the agent took in the target task $M_{target}$ to achieve threshold performance $\delta$, and $a^{M_i}$ is  the  number  of  actions  the  agent took in the source task $M_i$ of the curriculum.

Our objective is to find a curriculum $T = \{\mathcal{V}, \mathcal{E}, g, \mathcal{M}\}$ that improves an agent's time-to-threshold performance on the target task, given the LTL$_f$ representation and the OOMDP and MDP description of the target task. 




\section{Methodology}


The high-level curriculum generation algorithm is given in Algorithm~\ref{alg:agcg}. To begin, we assume the following:
\begin{enumerate}
    \item We have access to a high-level target task ($M_{tar}$) objective which is expressed using LTL$_f$ formulas, and can be equivalently represented using a DFA $\mathcal{U}$.
    \item  We have access to the OOMDP description of the target task. The OOMDP description contains the set of classes $\mathcal{C}$ of the environment, and for each class, the set of parameters for that class, along with the range of acceptable values for the parameters. Varying these parameters give us different tasks of the curriculum. We also have the initial and goal OOMDP states of the final task of the environment. 
    \item We have access to the functions $P$ that given a node in the DFA, return the set of MDPs that can reach the node, and access to  function $w$ that performs mapping between MDP states and OOMDP states.
\end{enumerate}


Given the DFA, we first generate a key-value pair dictionary $K$, where each key is a node of the automaton, and the value is the set of MDPs that can reach the node, with each MDP augmented with the initial and final OOMDP state configuration of the MDP, resulting in the MDP-OOMDP tuple $(M, s_{0,oo}, s_{f,oo})$, lines 1-2. The function {\tt Get\_Trace\_Paths} returns the set of all paths $(Z)$ that start from the initial node $\omega_0$ and reach an accepting node $\omega_f \in F$. We consider only acyclic DFAs, ignoring paths that contain a cycle. A curriculum designed with this consideration will prune curriculum candidates that intend to learn the same task repeatedly, ensuring progress in the curriculum. Then, for each path $\zeta \in Z$, the {\tt List\_Candidates} function generates an ordered-list of sequence-based curriculum candidates by choosing a MDP-OOMDP tuple from each node of the path, and sequencing it according to the sequence of nodes in the path $\zeta$. We repeat this until we attain a set of all possible combinations of MDP-OOMDP tuple sequences (lines 4-7). This yields an exhaustive set $\Psi$ where each element in the set is an ordered list of MDP-OOMDP tuples, and the length of the list is equal to the number of transitions in the trace path. Thus, each list corresponds to a curriculum candidate $T$ where the vertices of the graph are the MDPs in the list, and the edges are defined by the sequence of the MDPs in the list.
Once we have a set of sequence-based curricula candidates, we choose an effective curriculum from this set. We introduce the jump score, where we assign a score for each pair of consecutive source tasks in the curriculum (lines 10-16).

\textbf{Jump score:} Inspired by the difficulty scores for supervised learning~\cite{weinshall2018curriculum}, we assign a jump score $J: (M_i, M_j) \rightarrow \mathbb{R} \in [0,1]$ for each pair of consecutive source tasks $M_i$ and $M_j$: 
\begin{multline}
\small
J_{M_i \rightarrow M_j} = 1/2*(sim_t (M_j, M_{tar}) - sim_t (M_{i}, M_{tar})\\ +  sim_g (M_j, M_{tar}) - sim_g (M_{i}, M_{tar}))        
\end{multline}
where, $sim_t (M_i, M_{tar})$ is the task configuration similarity between the task $M_i$ and the final target task $M_{tar}$\footnote{More on MDP similarity:~\cite{visus2021taxonomy}}, while $sim_g (M_i, M_{tar})$ is the goal state similarity. The jump score $J_{M_i \rightarrow M_j}$ calculates how dissimilar two consecutive source tasks in the curriculum are by calculating the difference of the similarity of the task and goal configurations of each of the two consecutive tasks with the final target task. 

Since we have the OOMDP initial state configuration $s_{0,oo}$ of each pair of consecutive source tasks of the curriculum ($M_i, M_j$) and the final target task $M_{tar}$, the task configuration similarity between a task and the final target task is measured as the parameter value overlap between the initial OOMDP state values of the two tasks.
\begin{equation}
\small
sim_t(M_i, M_{tar}) = \frac{1}{\sum_{o \in \mathcal{O}_{M_{tar}}} |o.p^0|} \sum_{o \in \mathcal{O}_{M_{tar}}} \sum_{o.p \in o} \frac{o.p^0_{M_i}}{o.p^0_{M_{tar}}} 
\end{equation}
where $o.p^0_{M_i}$ is the value of the parameter of the object $o$ in the OOMDP state $s_{0,oo}$ of task $M_i$.
Likewise, we also have the OOMDP goal state $s_{f,oo}$ of each pair of consecutive source tasks and the final target task. We calculate the OOMDP goal state similarity of a task with respect to the final target task using:
\begin{equation}
\small
    sim_g(M_i, M_{tar}) = \frac{1}{\sum_{o \in \mathcal{O}_{M_{tar}}} |o.p^f|} \sum_{o \in \mathcal{O}_{M_{tar}}} \sum_{o.p \in o} \frac{o.p^f_{M_i}}{o.p^f_{M_{tar}}}     
\end{equation}
where $o.p^f_{M_i}$ is the value of the parameter of the object $o$ in the OOMDP state $s_{f,oo}$ of task $M_i$.

A lower jump score corresponds to a pair of tasks that have similar initial OOMDP task configurations as well as similar OOMDP goal state configurations. Along any path $\zeta$, the sum of the jump scores among all consecutive task pairs will always be equal to $1$. The set of OOMDP states the agent visits in an episode can be potentially large, and do not yield much information that might help in choosing a suitable curriculum candidiate. Additionally, access to the set of OOMDP states that the agent might visit in an episode before the agent even attempts the episode requires knowledge of the transition dynamics, which is unavailable. Hence, only the initial and final states are used while calculating the jump score.

Next, we calculate the average jump score of each curriculum candidate $\psi \in \Psi$ and store it in a Curriculum - Jump Score dictionary $\mathcal{J}$ (line 16). For sequence-based curricula, the curriculum is determined by the candidate that yields the lowest average jump score (lines 17-18). 
\begin{equation} 
    T = \operatorname*{argmin}_{\psi \in \Psi}  \mathcal{J}(\overline{J}_{\psi})
\end{equation}
Intuitively, the curriculum $T$ is an ordered list of MDP-OOMDP tuples that have the lowest cumulative average jump among any two consecutive source tasks, i.e. the curriculum has the lowest task and goal state dissimilarity between any two consecutive tasks. A lower dissimilarity will result in a stronger knowledge transfer.

The graph-based curriculum is determined by the set of elements in $\Psi$ that yield the cumulative average jump score lower than a predetermined threshold value $\eta$ (lines 19-21).
\begin{equation} 
    T' = \{\psi \in \Psi \ni \mathcal{J}(\overline{J}_{\psi}) \leq \eta \}
\end{equation}
Thus, $T'$ is a set where each element is an ordered list of MDPs (each element is a sequence-based curriculum). A graph is generated by considering equivalent MDPs in $T'$ as common nodes, and the edges are given by the sequence of MDPs in each element in $T'$ (line 22). The weights on the edges denote the proportion of knowledge contribution of a source task and is discussed further below. The worst case time-complexity of the algorithm is $O(|\mathcal{V}|^3|\mathcal{E}||C|^2|p|^2)$ and where $|\mathcal{V}|$ and $|\mathcal{E}|$ are the number of vertices and edges in the DFA’s DAG and $|C|$ and $|p|$ are the number of classes and maximum number of class parameters in the OOMDP. 

\textbf{Knowledge Transfer} in curriculum learning involves leveraging learned knowledge from a source task and transferring relevant knowledge to the next task in the curriculum. In our curriculum setting, each individual source task is learned using DQN~\cite{mnih_human-level_2015}, and hence we perform value function transfer, where the weights of the neural network of the learned value function of a source task are initialized as the weights of the value function of the next task, i.e. $V^i_{M_j} \leftarrow V^*_{M_i}$. Thus, instead of commencing the next task $M_j$ with random $V(s)$ values, the learned value function of the source task biases action selection in the next task according to the experience already collected in the source task. In scenarios where the action spaces are continuous, the RL algorithm needs to be adapted to suit the needs of the domain. DDPG~\cite{lillicrap2016continuous} and PPO~\cite{DBLP:journals/corr/SchulmanWDRK17} allow continuous action spaces and can replace the DQN as the base RL algorithm. In case of a sequence-based curriculum, where we transfer knowledge from only one source task $M_i$ to the next task $M_j$ in the curriculum, the value function of $M_j$ is initialized to be the learned value function of $M_i$. This process is continued until we reach the end of curriculum, culminating in the final target task $M_{tar}$.  
In case of a graph-based curriculum, where we transfer knowledge from multiple source tasks ($M_i, M_{i+1}, \ldots$) to the next task in the curriculum ($M_j$), the value function of the next task ($M_j$) is initialized to be a weighted sum of the value functions of its source tasks, i.e.
\begin{equation} \label{eq:1}
    V^i_j = \beta_0 \cdot V^*_{i} + \beta_1 \cdot V^*_{i+1} + \ldots +  \beta_l \cdot V^*_{i+l}
\end{equation}
where $V^i_j$ is the initial value function of the target task $M_j$, $V^*_{i}$ is the learned value function of the source task $M_i$, $l$ is the number of source tasks, and $\beta$ is a scalar weighting factor. To calculate $\beta$, we use the following:
\begin{equation}\label{eq:2}
    \beta_i \propto \frac{1}{J_{M_i \rightarrow M_j}} \: \: \: 
    \mbox{and} \: \: \:
    \sum_{k=0}^{k=l} \beta_{i+k} = 1
\end{equation}
i.e., the lower the jump score between tasks $M_i$ and $M_j$, the higher the similarity between their OOMDP task and goal configurations, thus, higher the weightage of the learned value function of the task $M_i$, and vice versa. A higher similarity will correspond to a stronger positive transfer. In case of a graph-based curriculum, we learn the tasks that correspond to the leaf nodes initially, transferring knowledge using Equations~\ref{eq:1} and ~\ref{eq:2}, culminating in the final target task\footnote{Link to source code and appendix: \\ https://github.com/tufts-ai-robotics-group/Automaton-guided-CL}. 

\begin{algorithm}[t]
\small
\caption{{\tt AGCG($\mathcal{U}, M_{tar}, w, P, \eta,\!$ CL-Type)}}
\label{alg:agcg}
\raggedright \textbf{Output}: Curriculum: $T$\\
\textbf{Placeholder Initialization}:
Set of all acyclic trace paths $Z \! \leftarrow \!\emptyset$ \\ 
Node-MDP Dictionary $K \leftarrow \emptyset$\\
Curriculum - Jump Score dictionary $\mathcal{J} \leftarrow \emptyset$\\
Set of curriculum candidates $\Psi \leftarrow \emptyset$\\
\textbf{Algorithm:}
\begin{algorithmic}[1] 
\FOR{$\omega \in \Omega$} 
\STATE $K \leftarrow K \cup \{\omega : (P(\omega), w(P(\omega)(\mathcal{S}_0)), w(P(\omega)(\mathcal{S}_f)) \}$
\ENDFOR
\STATE $Z \leftarrow $ {\tt Get\_Trace\_Paths}$(\mathcal{U})$
\FOR{$\zeta \in Z$}
\FOR{$\omega_i \in Occ(\zeta)$}
\STATE $\Psi \leftarrow \Psi \cup \{[${\tt List\_Candidates}$(K(\omega_i))]\}$
\ENDFOR
\ENDFOR
\FOR{$\psi \in \Psi$}
\STATE $jump\_sum \leftarrow 0$
\FOR{$i = 1 \: \mbox{to} \: |\psi|-1$}
\STATE $jump\_sum \leftarrow jump\_sum + J_{M_i\rightarrow M_{i+1}}$
\ENDFOR
\STATE $\overline{J}_{\psi} \leftarrow jump\_sum / |\psi|$
\STATE $\mathcal{J} \leftarrow \mathcal{J} \cup \{\psi: \overline{J}_{\psi} \}$
\ENDFOR
\IF{CL-Type == Sequence-based}
\STATE $T = \operatorname*{argmin}_{\psi \in \mathcal{J}}  \mathcal{J}(\overline{J}_{\psi})$

\ELSIF{CL-Type == Graph-Based} 
\STATE $T' = \{\psi \in \Psi \ni \mathcal{J}(\overline{J}_{\psi}) \leq \eta \}$
\STATE $T \leftarrow${\tt Graph}$(T')$
\ENDIF
\STATE \textbf{return} $T$
\end{algorithmic}
\end{algorithm}

\section{Experimental Results}
We aim to answer the following questions: (1) Does AGCL yield sample efficient learning? (2) How does it perform in environments that have distractor objects that are not modeled in the LTL$_f$ specification? (3) How does it perform when the exact OOMDP description is unknown? (4) Does it yield sample efficient learning when the OOMDP parameter space is continuous? (5) How does it perform when generating all potential curricula candidates is intractable?


\subsection{AGCL - Gridworld Results}\label{res:gridworld}

To answer our first question, we evaluated \emph{AGCL} on a grid-world domain with the LTL$_f$ objective:
\begin{multline}
    \textbf{G}((t \rightarrow \neg r \wedge \neg p) \wedge (r \rightarrow \neg t \wedge \neg p)\wedge (p \rightarrow \neg r \wedge \neg t))\\\wedge (\neg p\textbf{U}(t \wedge \textbf{X}(\neg p \textbf{U}t))) \wedge (\neg p\textbf{U}r)\wedge \textbf{F}p    
\end{multline}
where $t, r, p$ correspond to the atomic propositions $tree, rock$ and $pogo$-$stick$ respectively. Essentially, the agent needs to collect $2$ \textit{trees} and $1$ \textit{rock} (by navigating to the objects and \textit{breaking} them) before approaching the crafting table to \textit{craft} a \textit{pogo-stick}. This is a complex sequential decision making task and requires $\sim\!10^8$ interactions to reach convergence~\cite{shukla2022acute}. The OOMDP description is given by: $\mathcal{C} = \{world\_size, trees, rocks, crafting\_table\}$, where the class parameters are: $C(world\_size) = \{width, height\}$; $C(x) = \{x_{env}, x_{inv} | x \in \{trees,rocks \} \}$ and $C(crafting\_table) = \{crafting\_table_{env}\}$. The parameters $width$ and $height$ can assume values in the range $[6,12]$ and the number of $trees, rocks$ and $crafting\_table$ in the environment can assume a value in the ranges of $[0,4]$, $[0,2]$ and $[0,1]$ respectively. In this environment, the agent can move 1 cell forward if the cell ahead is clear or rotate $\pi/2$ left or right. In the target task, the agent receives a reward of $10^3$ upon crafting a \textit{pogo-stick}, and $-1$ reward for all other steps.  The agent's sensor emits a beam at incremental angles of $\pi/4$ to determine the closest object in the angle of the beam (i.e., the agent receives a local-view of its environment). Two additional sensors provide information on the amount of \textit{trees} and \textit{rocks} in the agent's inventory (More details in Appendix B).

\begin{figure*}[ht]
	\centering
	\begin{minipage}{.24\textwidth}
		\centering
		\includegraphics[width=\textwidth,height=0.75\textwidth]{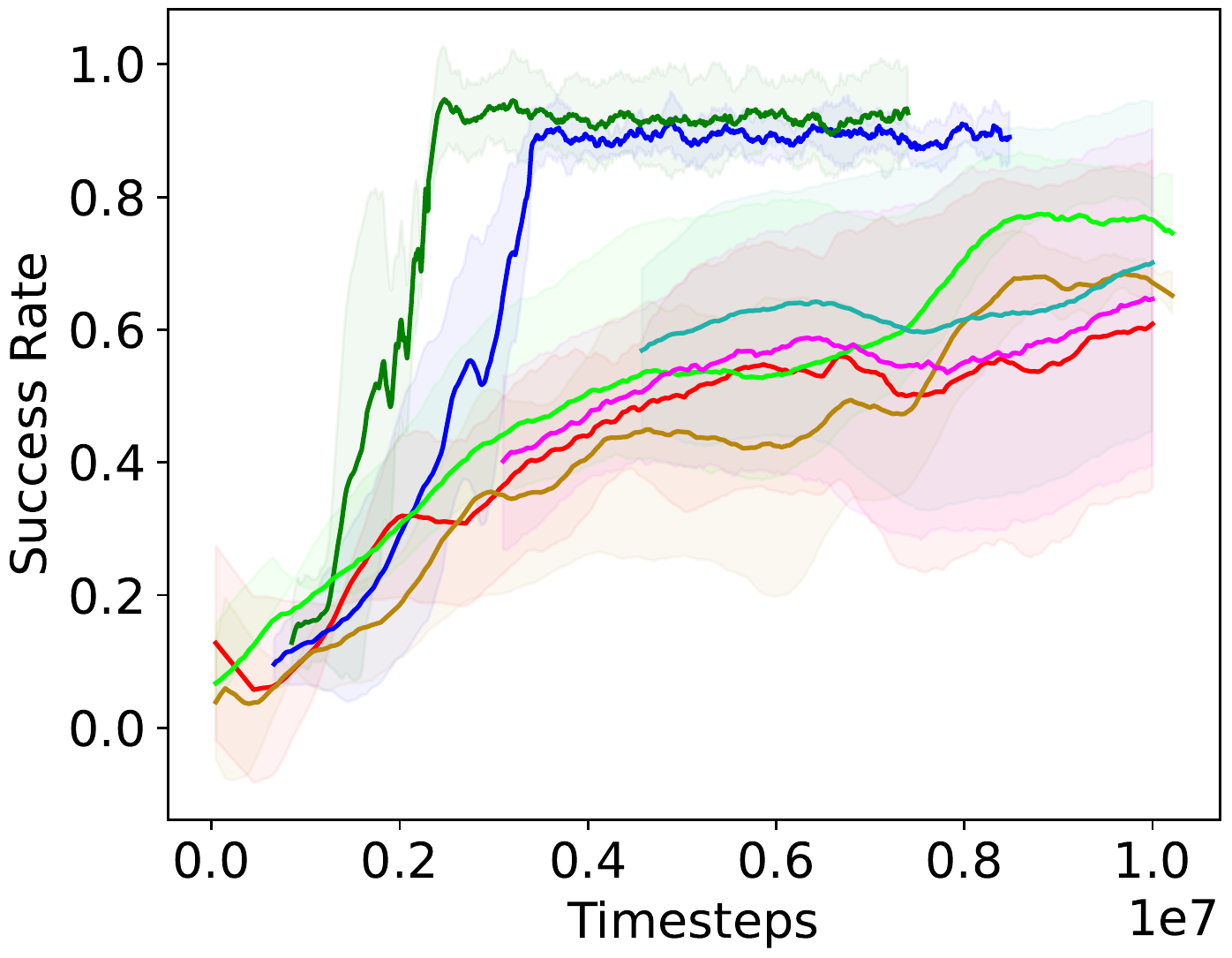}
        \subcaption{ Gridworld domain}
        \label{fig:MC}
	\end{minipage}%
	\hfill
	\centering
	\begin{minipage}{.245\textwidth}
		\centering
		\includegraphics[width=\textwidth,height=0.75\textwidth]{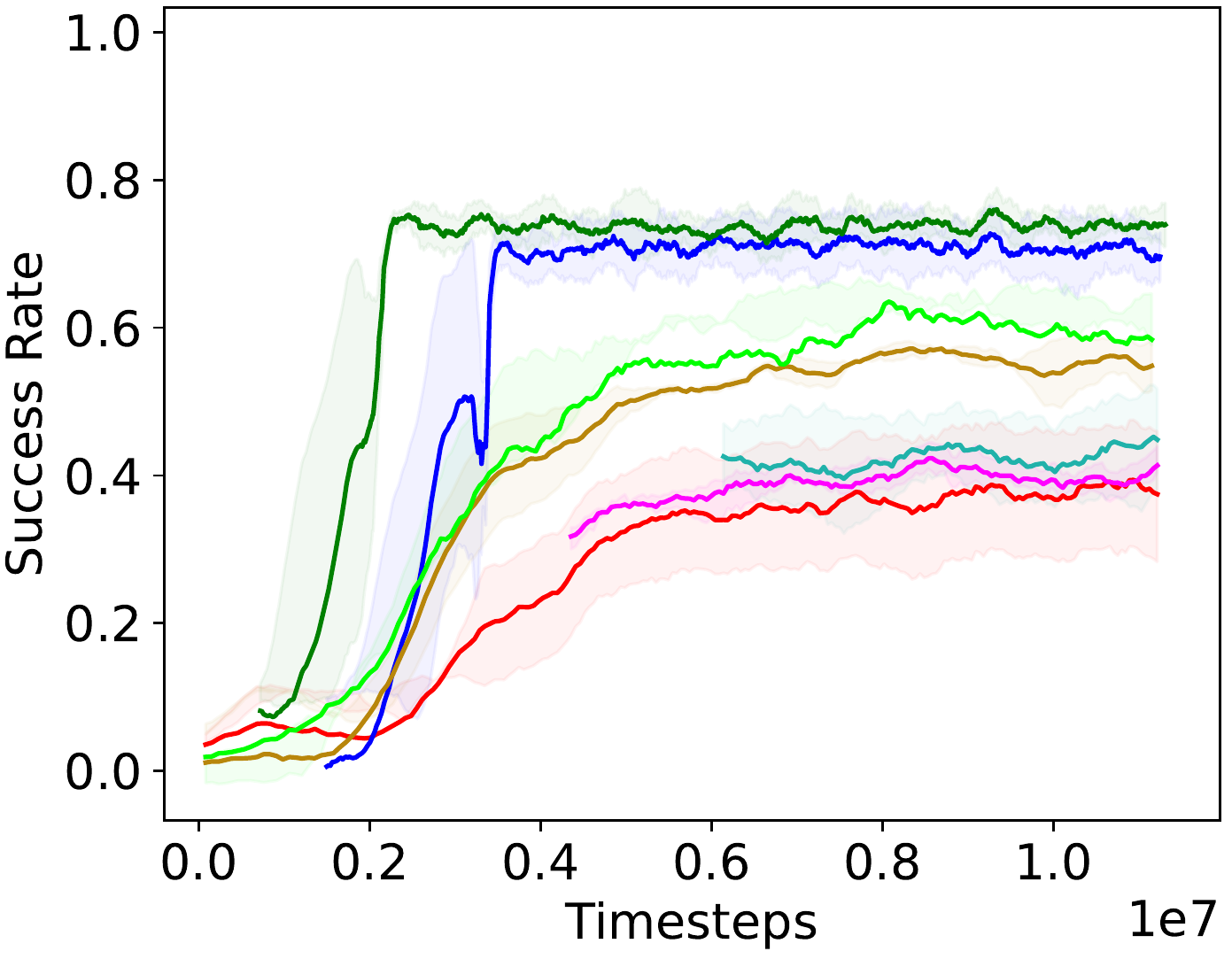}
        \subcaption{Distractor objects}
        \label{fig:MC-distractor}
	\end{minipage}%
	\hfill
	\centering
	\begin{minipage}{.245\textwidth}
		\centering
		\includegraphics[width=\textwidth,height=0.75\textwidth]{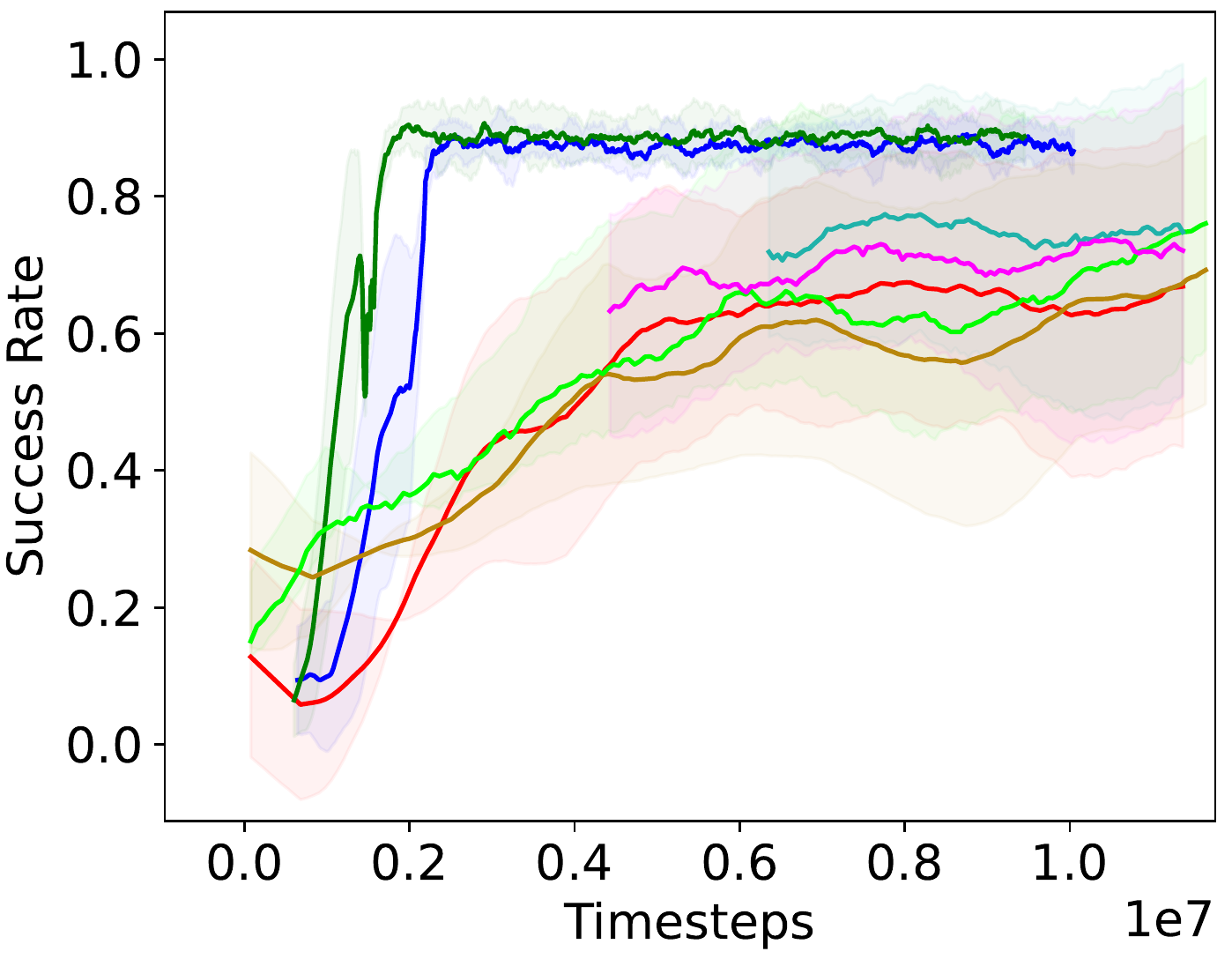}
        \subcaption{Imperfect OOMDP}
        \label{fig:MC-noisy}
	\end{minipage}%
	\hfill
	\centering
	\begin{minipage}{.245\textwidth}
		\centering
		\includegraphics[width=\textwidth,height=0.75\textwidth]{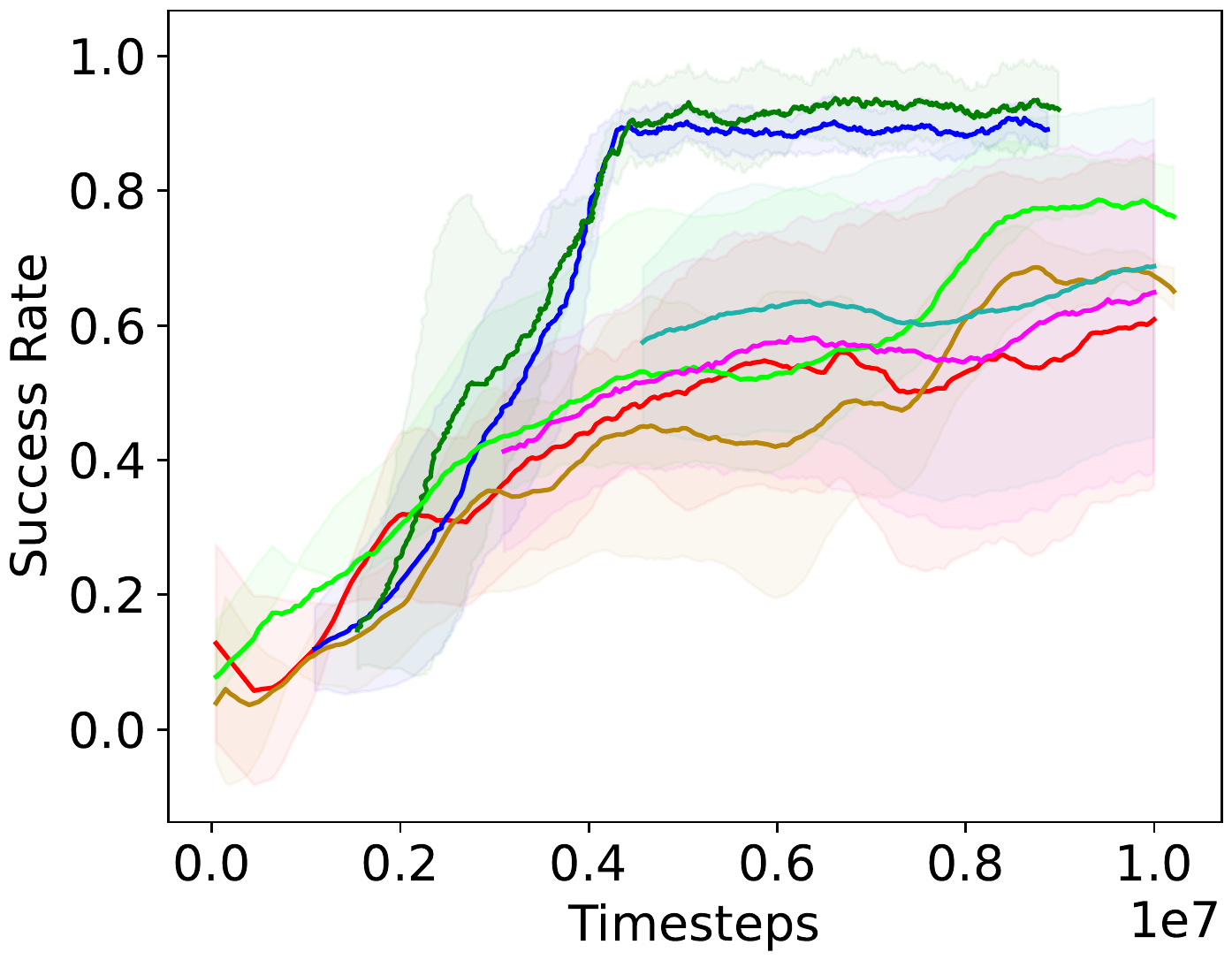}
        \subcaption{Subset candidates}
        \label{fig:MC-subset}
	\end{minipage}%
	\hfill	
 \vspace{1em}
	\begin{minipage}{0.245\textwidth}
		\centering
		\includegraphics[width=0.75\textwidth]{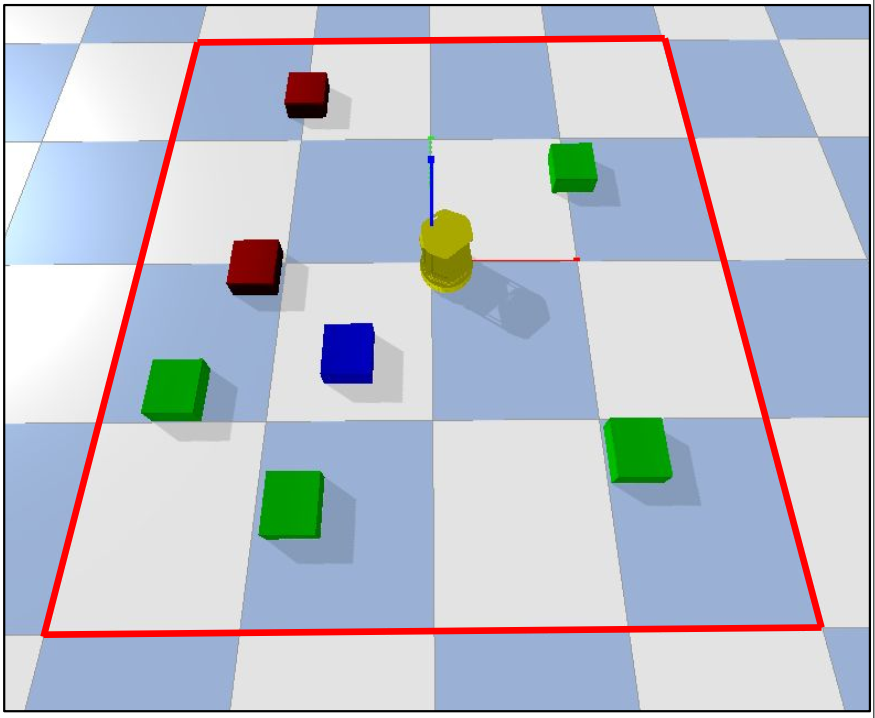}
        \subcaption{Crafter-Turtlebot domain}
        \label{fig:CT-domain}
	\end{minipage}%
	\hfill
	\centering
	\begin{minipage}{.245\textwidth}
		\centering
		\includegraphics[width=\textwidth,height=0.65\textwidth]{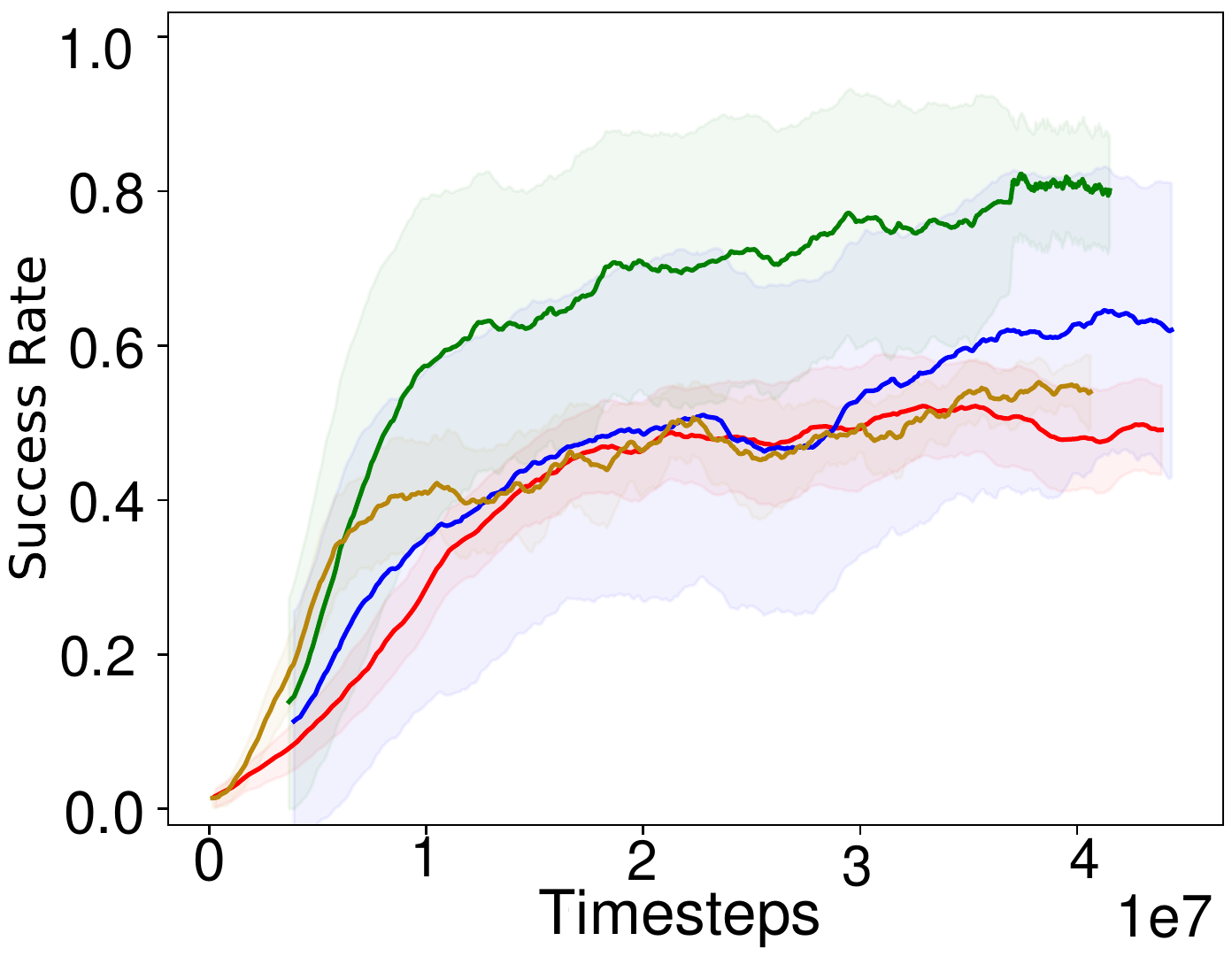}
        \subcaption{Crafter-Turtlebot}
        \label{fig:CT-results}
	\end{minipage}%
	\hfill
	\centering
	\begin{minipage}{.245\textwidth}
		\centering
		\includegraphics[width=0.7\textwidth,height=0.65\textwidth]{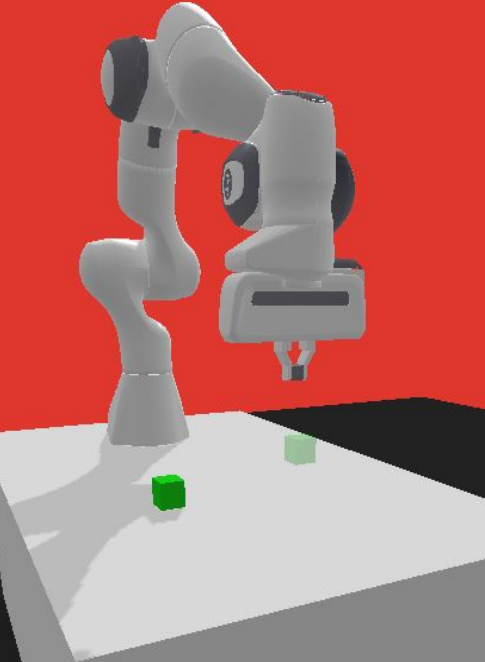}
        \subcaption{Panda-pick-and-place}
        \label{fig:Panda-domain}
	\end{minipage}%
	\centering
	\begin{minipage}{.245\textwidth}
		\centering
		\includegraphics[width=\textwidth,height=0.65\textwidth]{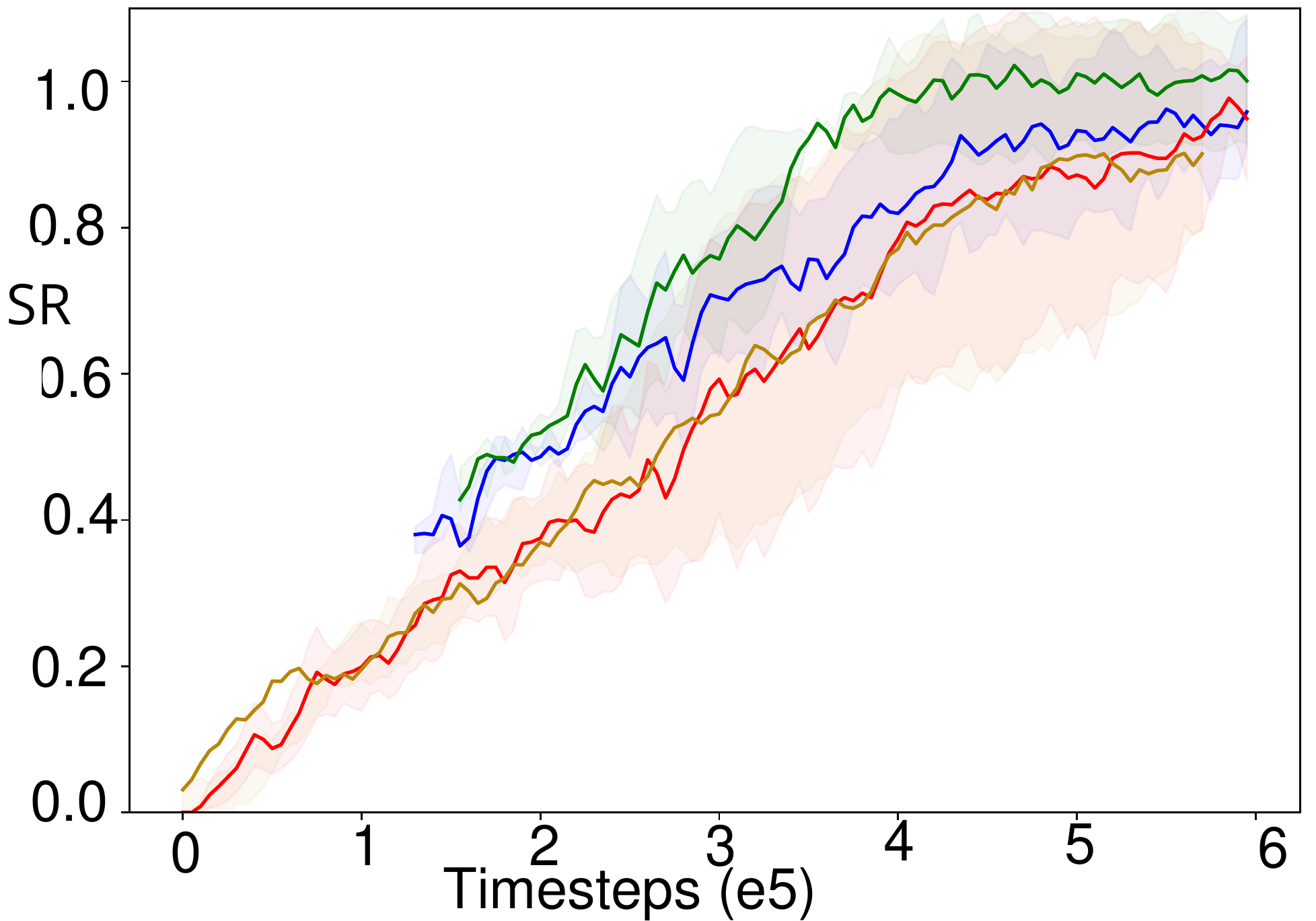}
        \subcaption{Panda-pick-and-place}
        \label{fig:Panda-results}
	\end{minipage}%
	\hfill	
\includegraphics[width = 0.9\textwidth, height=0.065\textwidth]{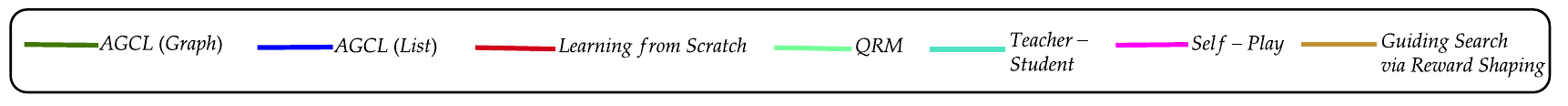}	
	\caption{ Learning curves (Averaged over 10 trials): (a) Gridworld domain; (b) Gridworld - with distractor objects in the environment; (c) Gridworld - with imperfect OOMDP description of the environment; (d) Gridworld - with subset of curriculum candidates. Figs.(e) and (g) show the two robotic environments and their learning curves in (f) and (h)}\label{fig:results1}
\end{figure*}

The learning curves in Fig.~\ref{fig:results1} depict the performance of our proposed sequence and graph-based \emph{AGCL} method against five baseline approaches namely learning from scratch, automaton-guided reward shaping baslines: GSRS~\cite{camacho2018non}, QRM~\cite{icarte2018using}, and curriculum learning baselines: Teacher-Student~\cite{DBLP:journals/tnn/MatiisenOCS20} and self-play~\cite{sukhbaatar2018intrinsic}, all implemented using the same RL algorithm DQN~\cite{mnih_human-level_2015}. The automaton-guided reward shaping baselines do not employ a curriculum and build upon naive reward shaping by modifying the reward inversely proportional to the distance from the DFA goal state and by learning policies for each DFA state transition. The curriculum learning baseline approaches do not utilize a reward machine, and rely on optimizing the task sequence through agent's experience\footnote{Baseline implementation details in Appendix Section C},.  
The learning curve for the curriculum approaches have an offset on the x-axis to account for the interactions used to go through the curriculum before moving on to the target task, signifying \textit{strong transfer} \cite{taylor2009transfer}. The results in Fig.~\ref{fig:MC} show that \emph{AGCL} reaches a successful policy quicker, and the graph-based curriculum proposed by $\emph{AGCL}$ outperforms the sequence-based approach. Learning source tasks in a graph-based curriculum is also parallelized, reducing the overall wall time. Furthermore, our method outperforms the five baseline approaches in terms of learning speed. The other four baseline approaches perform better than learning from scratch, but do not outperform \emph{AGCL}. All approaches may converge to a higher asymptote, but will need more interactions with the environment. For this task, the curriculum generated by \emph{AGCL} has $|\mathcal{V}| = 4$ and $|\mathcal{V}| = 6$ for the sequence-based and graph-based curricula respectively. 
To demonstrate that the average convergence rate of \emph{AGCL} is consistently higher than the baseline approaches, we perform an unpaired t-test~\cite{kim2015t} to compare \emph{AGCL} against the best performing baselines at the end of $10^7$ training interactions and we observed statistically significant results (95$\%$ confidence). Thus, \emph{AGCL} not only achieves a better success rate, but also converges faster.\footnote{Statistical significance result details in Appendix Section D.}

\subsection{AGCL on Tasks with Distractor Objects}
Next, we test our \emph{AGCL} approach (sequence and graph) on the \textit{pogo-stick} task described in Sec.~\ref{res:gridworld} but now the OOMDP description of the environment contains a new class $C(distractor)\! = \!\{distractor_{env}, distractor_{inv} \}$, whose object $distractor_{env}$ or $distractor_{inv}$ can assume values in the range $[0,2]$, signifying that the initial task state or the inventory may contain $0$ to $2$ instances of the $distractor$ object. The agent interacts with this object the same way it interacts with $trees$ and $rocks$, i.e. by navigating to the object and collecting it in inventory. However, presence of $distractor$ objects in the inventory of the agent is not necessary to achieve the target objectives and is not modeled in the LTL$_f$ specification.

Fig.~\ref{fig:MC-distractor} demonstrates how our proposed \emph{AGCL} performs against the baseline approaches. Since our curriculum generation approach models not only the goal configuration but also the initial task space configuration, it is successful in estimating the difficulty posed by the introduction of the distractor object as they are present in the initial state of the target task. The curriculum generated by \emph{AGCL} guides the agent to a successful policy in fewer interactions compared to the other baselines. Approaches that design a curriculum purely from the LTL$_f$ specification will not model the distractor objects, and will thus fail to determine the complexity of a source task that contains these distractor objects.

\subsection{Tasks with Imperfect OOMDP Descriptions}
In certain partially observable settings, it might not be possible to obtain an accurate OOMDP description of the environment. To demonstrate the efficacy of our approach in imperfect OOMDP descriptions, we incorporate a gaussian noise over the class parameter value ranges, i.e. if the parameter range is $Par_{given}(world\_size[height]) = [a,b]$, we assume that this range is imperfect, and the true range is given by $Par_{noisy}(world\_size[height]) = [a-\mathcal{N}(0, \sigma), b+\mathcal{N}(0, \sigma)]$, where $\sigma = (b-a)/6$; covering the entire range of the parameter values in six standard deviations. We incorporate a gaussian noise over all the class parameters ranges.


We test \emph{AGCL} (sequence and graph) on the same task of Sec.~\ref{res:gridworld} but now the environment's OOMDP description is noisy over the parameter values, i.e., the exact OOMDP description is unknown. We observe that even with imperfect OOMDP descriptions, \emph{AGCL} converges faster than the baselines (Fig~\ref{fig:MC-noisy}), and the graph-based \emph{AGCL} converges to a successful policy the quickest, in $2.5\times10^6$ interactions.

\subsection{Tasks with Continuous OOMDP Ranges}
In the gridworld, the OOMDP parameter values were integers. In this experiment, we test how our approach performs in tasks that have continuous parameter ranges. We modify the function $P: \Omega \rightarrow \mathcal{M}$ that given a node in the DFA, samples a subset of MDPs $\mathcal{M}_{sub} \subseteq \mathcal{M}$ such that $|\mathcal{M}_{sub}| = b$. We test this on two challenging simulated robotic environments where the interaction cost is high. 

Fig.~\ref{fig:CT-domain} shows a robotic task with the same objective described in Sec.~\ref{res:gridworld}. Here the values for the parameters $width,height$ for class $world\_size$ are in the continuous range $[2m, 4m]$. The move forward (backward) action causes the robot to move forward (backward) by $0.1m$ and the robot rotates by $\pi/8$ radians with each rotate action. Additionally, the objects can be placed at continuous locations in this robotic domain as compared to discrete grid locations for the gridworld task. These changes increase the number of MDP and OOMDP states the agent can attain.

The second simulated robotics environment (Fig.~\ref{fig:Panda-domain}) consists of a robotic arm performing a pick-and-place task~\cite{gallouedec2021pandagym} with the LTL$_f$ objective: \textbf{F}$(g \wedge $\textbf{F}$(p \wedge $\textbf{F}$q))$ where $g, p, q$ are the atomic propositions for \textit{'reach-object'}, \textit{`pick-up-object'} and \textit{`place-object'}. The OOMDP is modeled as $\mathcal{C}=$ $\{world\_size, objects\}$ with $Par(world\_size) =$ $\{length, width, height\}$ and $Par(objects) = \{no\_objects\}$. The continuous parameters $\{length, width, height\}$ assume values between $[10,50]cm$. The robot has continuous action parameters for moving the arm and a binary gripper action (close/open). The robotic domains were modeled using PyBullet~\cite{coumans2021}. For each node, we sample $b=25$ different MDPs and evaluate \emph{AGCL}. The results in Figs.(\ref{fig:CT-results},\ref{fig:Panda-results}), show that \emph{AGCL} outperforms all other baselines in both environments. Additionally, the graph-based \emph{AGCL} converges to a successful policy the quickest.


\subsection{AGCL on Subset of Curriculum Candidates}
In cases where the range of parameter values or the DFA representation is too large, the number of curriculum candidates can grow exponentially in terms of the number of nodes in the DFA, making it infeasible to calculate the jump-score for each curriculum candidate. For this, we evaluate \emph{AGCL} on the task described in Sec.~\ref{res:gridworld} by sampling a subset $\Psi_{sub}$ of the total curriculum candidates $\Psi$, such that $|\Psi_{sub}| = |\Psi|/4$, thus reducing the total number of curriculum candidates and the amount of computation required to determine a suitable curriculum. From the results in Fig.~\ref{fig:MC-subset}, we observe that in this experiment, \emph{AGCL} takes more interactions to converge to a desired policy as compared to Fig.~\ref{fig:MC}, as sampling a subset prunes some desirable curriculum candidates, however it outperforms the other curriculum learning and automaton-guided RL baselines in terms of learning speed.
 
\section{Conclusion and Future Work}

We proposed \emph{AGCL}, a framework for curriculum generation using the high-level LTL$_f$ specification and the OOMDP description of the environment. \emph{AGCL} decomposes the target objective into sub-goals, and generates a sequence-based or a graph-based curriculum for the task. Through experiments, we demonstrated that \emph{AGCL} accelerates learning, converging to a desired success rate quicker as compared to other curriculum learning and automaton-guided RL baselines. Moreover, our proposed approach improves learning performance even in the presence of distractor objects in the environment that are not modeled in the LTL$_f$ specification. Finally, we evaluate our approach on long-horizon complex robotic tasks where the state space is large. \emph{AGCL} reduces training time without relying on human-guided dense reward function nor does it require a perfect OOMDP description of the environment. The graph-based curricula produced by \emph{AGCL} affords learning separate behaviors in parallel, reducing wall clock time for complex sequential decision making tasks. Thus, our proposed \emph{AGCL} approach offers accelerated learning when the high-level task objective is available.

We assume that the high-level task objective is available, and can be characterized using a LTL$_f$ formula. However, in certain cases, the target tasks can be completely novel, with no prior information on the task objective. Additionally, it might not be feasible to represent the domain using OOMDP representation, as the objects in the environment can be unknown, and the only way to get more information about the environment is through interaction with the environment. In scenarios where the LTL$_f$ specification becomes complex, the DFA gets larger, and it might become intractable to come up with exact solutions of curriculum optimization. Future work would involve devising approximate solutions that tightly bound exact solutions. Also, our heuristic, the jump score, while it was shown to be useful here, might not perform universally well for all task objectives and environments. In future work, we plan to modify this heuristic to be more theoretically-grounded. Furthermore, we plan to extend it to settings where obtaining an accurate LTL$_f$ specification might be difficult. 
Finally, we would like to extend our work to multi-agent settings, and investigate curriculum generation using even more expressive high-level languages, such as CTL$^*$ and $\mu$-calculus.

\section*{Ethical Impact}

\emph{AGCL} aims to boost the learning speed of reinforcement learning agents for sequential decision making tasks. This will enable quicker learning for robotic applications. It will lead to an overall reduction in the training times, saving computation time and energy. On the other hand, this work can also be used for negative applications - e.g., malicious use of RL. However, the above mentioned concerns are central to all works that deal with any aspects of AI.

\section*{Acknowledgements}
A portion of this work was conducted in the Multimodal Learning, Interaction, and Perception Lab at Tufts University, Assured Information Security, Inc., Georgia Tech Research Institute, and the University of Colorado Boulder, with support from the Air Force Research Lab under contract FA8750-22-C-0501.


\bibliography{aaai23}

\end{document}